\documentclass[11pt,a4paper]{article}
\usepackage{authblk}
\usepackage[hyperref]{emnlp2018}
\usepackage{times}
\usepackage{latexsym}
\usepackage{amsmath}
\usepackage{amsfonts}
\usepackage{multirow}
\usepackage{comment}

\usepackage{fancyhdr}
\usepackage{lastpage}
\usepackage{url}

\usepackage{tikz}
\usepackage{booktabs}

\usepackage{bm}
\usepackage{relsize}
\usepackage{pgfplots}
 
\usetikzlibrary{arrows,shapes, decorations.pathmorphing,backgrounds,positioning}

\usepackage{caption} 
\usepackage{subcaption}

\definecolor{purple}{RGB}{102, 0, 255}
\definecolor{mygreen}{RGB}{45, 114, 70}
\aclfinalcopy 

\title{Pyramidal Recurrent Unit for Language Modeling}

\date{}

\author[1]{Sachin Mehta}
\author[1]{Rik Koncel-Kedziorski}
\author[2]{Mohammad Rastegari}
\author[1]{Hannaneh Hajishirzi}

\affil[1]{University of Washington, Seattle, WA, USA}
\affil[ ]{\{sacmehta, kedzior, hannaneh\}@uw.edu}
\affil[2]{Allen Institute for AI and XNOR.AI, Seattle, WA, USA}
\affil[ ]{mohammadr@allenai.org}

\begin{document}
\maketitle
\begin{abstract}
LSTMs are powerful tools for modeling contextual information, as evidenced by their success at the task of language modeling. However, modeling contexts in very high dimensional space can lead to poor generalizability. We introduce the Pyramidal Recurrent Unit (PRU), which enables learning representations in high dimensional space with more generalization power and fewer parameters. PRUs replace the linear transformation in LSTMs with more sophisticated interactions including pyramidal and grouped linear transformations. This architecture gives strong results on word-level language modeling while reducing the number of parameters significantly. In particular, PRU improves the perplexity of a recent state-of-the-art language model \citet{merity2017regularizing} by up to 1.3 points while learning 15-20\% fewer parameters. For similar number of model parameters, PRU outperforms all previous RNN models that exploit different gating mechanisms and transformations. We provide a detailed examination of the PRU and its behavior on the language modeling tasks. Our code is open-source and available at \url{https://sacmehta.github.io/PRU/}.
\end{abstract}

\section{Introduction}
\begin{figure}[t!]
\centering
\includegraphics[width=\columnwidth]{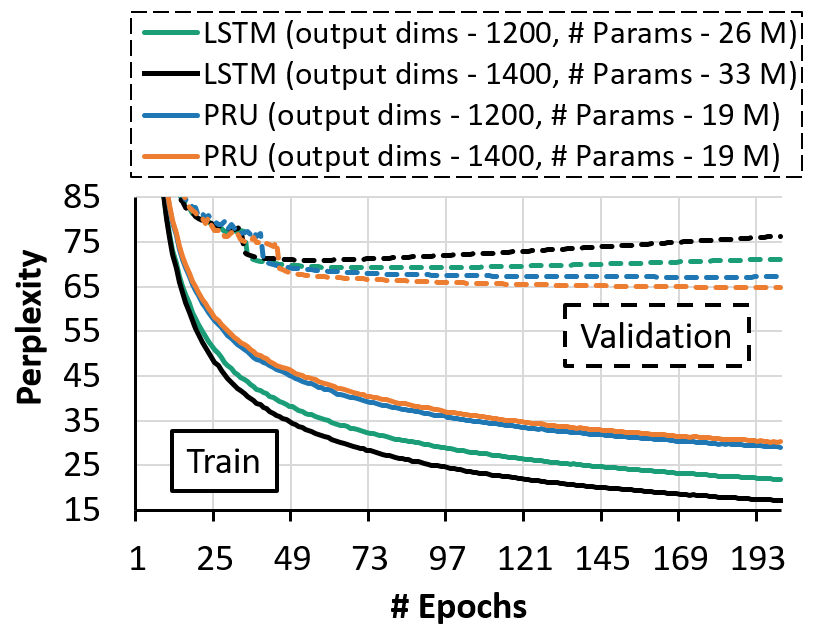}
\caption{Comparison of training (solid lines) and validation (dashed lines) perplexities on the Penn Treebank with standard dropout for pyramidal recurrent units (PRU) and LSTM. PRUs learn latent representations in very high-dimensional space with good generalizability and fewer parameters. See Section \ref{sec:pruArch} for more details about PRUs. Best viewed in color.}
\label{fig:lstmOverHD}
\end{figure}

\begin{figure*}[t!]
\centering
\includegraphics[width=2\columnwidth]{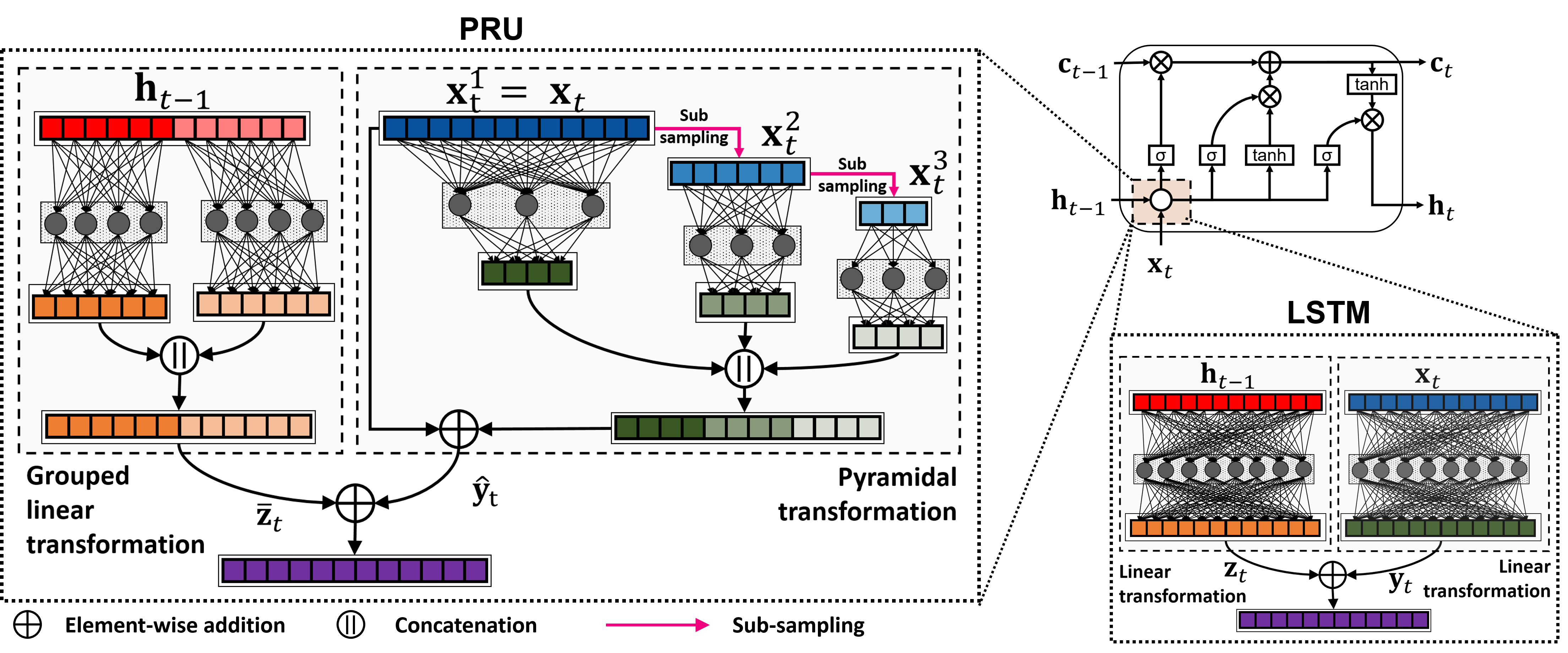}
\caption{Block diagram visualizing the transformations in pyramidal recurrent unit (left) and the LSTM (bottom right) along with the LSTM gating architecture (top right). \textcolor{blue}{\bf Blue}, \textcolor{red}{\bf red}, \textcolor{mygreen}{\bf green} (or \textcolor{orange}{\bf orange}), and \textcolor{purple}{\bf purple} signify the current input $\mathbf{x}_t$, output of the previous cell $\mathbf{h}_{t-1}$, the output of transformations, and the fused output, respectively. The color intensity is used to represent sub-sampling and grouping operations.}
\label{fig:lstmvsPRU}
\end{figure*}

Long short term memory (LSTM) units \cite{hochreiter1997long} are popular for many sequence modeling tasks and are used extensively in language modeling. A key to their success is their articulated gating structure, which allows for more control over the information passed along the recurrence. However, despite the sophistication of the gating mechanisms employed in LSTMs and similar recurrent units, the input and context vectors are treated with simple linear transformations prior to gating. Non-linear transformations such as convolutions \cite{kim2016character} have been used, but these have not achieved the performance of well regularized LSTMs for language modeling \cite{melis2017state}. 

A natural way to improve the expressiveness of linear transformations is to increase the number of dimensions of the input and context vectors, but this comes with a significant increase in the number of parameters which may limit generalizability. An example is shown in Figure \ref{fig:lstmOverHD}, where LSTMs performance decreases with the increase in dimensions of the input and context vectors. Moreover, the semantics of the input and context vectors are different, suggesting that each may benefit from specialized treatment. 

Guided by these insights, we introduce a new recurrent unit, the Pyramidal Recurrent Unit (PRU), which is based on the LSTM gating structure. Figure \ref{fig:lstmvsPRU} provides an overview of the PRU. At the heart of the PRU is the {\it pyramidal transformation} (PT), which uses subsampling to effect multiple views of the input vector. The subsampled representations are combined in a pyramidal fusion structure, resulting in richer interactions between the individual dimensions of the input vector than is possible with a linear transformation. Context vectors, which have already undergone this transformation in the previous cell, are modified with a \textit{grouped linear transformation} (GLT) which allows the network to learn latent representations in high dimensional space with fewer parameters and better generalizability (see Figure~\ref{fig:lstmOverHD}).

We show that PRUs can better model contextual information and demonstrate performance gains on the task of language modeling. The PRU improves the perplexity of the current state-of-the-art language model \cite{merity2017regularizing} by up to 1.3 points, reaching perplexities of 56.56 and 64.53 on the Penn Treebank  and WikiText2 datasets while learning 15-20\% fewer parameters. Replacing an LSTM with a PRU results in improvements in perplexity across a variety of experimental settings. We provide detailed ablations which motivate the design of the PRU architecture, as well as detailed analysis of the effect of the PRU on other components of the language model. 

\section{Related work}
Multiple methods, including a variety of gating structures and transformations, have been proposed to improve the performance of recurrent neural networks (RNNs). We first describe these approaches and then provide an overview of recent work in language modeling.

\paragraph{Gating-based mechanisms:} The performance of RNNs have been greatly improved by gating mechanisms such as LSTMs \cite{hochreiter1997long}, GRUs \cite{chung2014empirical}, peep-hole connections \cite{gers2000recurrent}, SRUs \cite{lei2017sru}, and RANs \cite{lee2017recurrent}. In this paper, we extend the gating architecture of LSTMs \cite{hochreiter1997long}, a widely used recurrent unit across different domains.

\paragraph{Transformations:} Apart from the widely used linear transformation for modeling the temporal data, another transformation that has gained popularity is convolution \cite{lecun1995convolutional}. Convolution-based methods have gained attention in computer vision tasks \cite{krizhevsky2012imagenet} as well as some of the natural language processing tasks including machine translation \cite{gehring2017convolutional}. Convolution-based methods for language modeling, such as CharCNN \cite{kim2016character}, have not yet achieved the performance of well regularized LSTMs \cite{melis2017state}. We inherit ideas from convolution-based approaches, such as sub-sampling, to learn richer representations \cite{krizhevsky2012imagenet,han2017deep}.

\paragraph{Regularization:} Methods such as dropout \cite{srivastava2014dropout}, variational dropout \cite{kingma2015variational}, and weight dropout \cite{merity2017regularizing} have been proposed to regularize RNNs. These methods can be easily applied to PRUs.

\paragraph{Other efficient RNN networks:} Recently, there has been an effort to improve the efficiency of RNNs. These approaches include quantization \cite{Chen2018alter}, skimming \cite{seo2017neural,yu2017learning}, skipping \cite{campos2017skip}, and query reduction \cite{seo2016query}. These approaches extend standard RNNs and therefore, these approaches are complementary to our work.

\paragraph{Language modeling:} Language modeling is a fundamental task for NLP and has garnered significant attention in recent years (see Table \ref{tab:sota} for comparison with state-of-the-art methods). \citet{merity2017regularizing} introduce regularization techniques such as weight dropping which, coupled with a non-monotonically triggered ASGD optimization, achieves strong performance improvements.  \citet{yang2017breaking} extend \citet{merity2017regularizing} with the mixture of softmaxes (MoS) technique, which increases the rank of the matrix used to compute next-token probabilities. Further, \citet{merity2016pointer} and \citet{krause2017dynamic} propose methods to improve inference by adapting models to recent sequence history. Our work is complementary to these recent softmax layer and inference procedure improvements. 

We extend state-of-the-art language model in \citet{merity2017regularizing} by replacing the LSTM with the PRU. We show by experiments that the PRU improves the performance of \citet{merity2017regularizing} while learning fewer parameters. 

\section{Pyramidal Recurrent Units}
\label{sec:pruArch}
We introduce Pyramidal Recurrent Units (PRUs), a new RNN architecture which improves modeling of context by allowing for higher dimensional vector representations while learning fewer parameters. Figure \ref{fig:lstmvsPRU} provides an overview of PRU. We first elaborate on the details of the pyramidal transformation and the grouped linear transformation. We then describe our recurrent unit, PRU.

\subsection{Pyramidal transformation for input}
The basic transformation in many recurrent units is a linear transformation $\mathcal{F}_L$ defined as:
\begin{equation}
\mathbf{y} = \mathcal{F}_L(\mathbf{x}) = \mathbf{W} \cdot \mathbf{x}, 
\label{eq:lin}
\end{equation}
where $\mathbf{W} \in \mathbb{R}^{N\times M}$ are learned weights that linearly map $\mathbf{x} \in \mathbb{R}^N$ to $\mathbf{y} \in \mathbb{R}^M$. To simplify notation, we omit the biases.

Motivated by successful applications of sub-sampling in computer vision (e.g., \cite{burt1987laplacian, lowe1999object, krizhevsky2012imagenet, mehta2018espnet}), we subsample input vector $\mathbf{x}$ into $K$ pyramidal levels to achieve representation of the input vector at multiple scales. This sub-sampling operation produces $K$ vectors, represented as $\mathbf{x}^k \in \mathbb{R}^{\frac{N}{2^{k-1}}}$, where $2^{k-1}$ is the sampling rate and $k=\{1, \cdots, K\}$. We learn scale-specific transformations $\mathbf{W}^k \in \mathbb{R}^{\frac{N}{2^{k-1}} \times \frac{M}{K}}$ for each $k = \{1, \cdots K\}$. The transformed subsamples are concatenated to produce the pyramidal analog to $\mathbf{y}$, here denoted as $\mathbf{\bar y} \in \mathbb{R}^M$:
\begin{equation}
\mathbf{\bar y} = \mathcal{F}_P(\mathbf{x}) = \left[ \mathbf{W}^1 \cdot \mathbf{x}^1, \cdots, \mathbf{W}^K \cdot \mathbf{x}^K\right],
\label{eq:pyr}
\end{equation}
where $\left[\cdot, \cdot \right]$ indicates concatenation. We note that pyramidal transformation with $K=1$ is the same as the linear transformation.

To improve gradient flow inside the recurrent unit, we combine the input and output  using an element-wise sum (when dimension matches) to produce residual analog of pyramidal transformation, as shown in Figure~\ref{fig:lstmvsPRU} \cite{he2016deep}. 

\paragraph{Sub-sampling:} We sub-sample the input vector $\mathbf{x}$ into $K$ pyramidal levels using the kernel-based approach \cite{lecun1995convolutional, krizhevsky2012imagenet}. Let us assume that we have a kernel $\kappa$ with $2e + 1$ elements. Then, the input vector $\mathbf{x}$ can be sub-sampled as:
\begin{equation}
\mathbf{x}^k = \sum\limits_{i=1}^{N/s} \sum\limits_{j=-e}^{e} \mathbf{x}^{k-1}[si] \kappa[j],
\label{eq:kern}
\end{equation}
where $s$ represents the stride and $k=\{2,\cdots, K\}$.

\paragraph{Reduction in parameters:} The number of parameters learned by the linear transformation and the pyramidal transformation with $K$ pyramidal levels to map $\mathbf{x} \in \mathbb{R}^N$ to $\mathbf{\bar y} \in \mathbb{R}^M$ are $NM$ and $\frac{NM}{K} \sum\limits_{k=1}^K 2^{(1-k)}$ respectively. Thus, pyramidal transformation reduces the parameters of a linear transformation by a factor of $K(\sum_{k=1}^K 2^{(1-k)})^{-1}$. For example, the pyramidal transformation (with $K=4$ and $N=M=600$) learns $53\%$ fewer parameters than the linear transformation. 

\subsection{Grouped linear transformation for context}
Many RNN architectures apply linear transformations to both the input and context vector. However, this may not be ideal due to the differing semantics of each vector. In many NLP applications including language modeling, the input vector is a dense word embedding which is shared across all contexts for a given word in a dataset. In contrast, the context vector is highly contextualized by the current sequence. The differences between the input and context vector motivate their separate treatment in the PRU architecture. 

The weights learned using the linear transformation (Eq.~\ref{eq:lin}) are reused over multiple time steps, which makes them prone to over-fitting \cite{gal2016theoretically}. To combat over-fitting, various methods, such as variational dropout \cite{gal2016theoretically} and weight dropout \cite{merity2017regularizing}, have been proposed to regularize these recurrent connections. To further improve generalization abilities while simultaneously enabling the recurrent unit to learn representations at very high dimensional space, we propose to use grouped linear transformation (GLT) instead of standard linear transformation for recurrent connections \cite{kuchaiev2017factorization}. While pyramidal and linear transformations can be applied to transform context vectors, our experimental results in Section~\ref{sec:abl} suggests that GLTs are more effective.

The linear transformation $\mathcal{F}_L : \mathbb{R}^N \to \mathbb{R}^M$ maps $\mathbf{h} \in \mathbb{R}^N$ linearly to $\mathbf{z} \in \mathbb{R}^M$. Grouped linear transformations break the linear interactions by factoring the linear transformation into two steps. First, a GLT splits the input vector $\mathbf{h} \in \mathbb{R}^N$ into $g$ smaller groups such that $\mathbf{h} = \{\mathbf{h}^1, \cdots, \mathbf{h}^g\}, \forall\ \mathbf{h}^i \in \mathbb{R}^{\frac{N}{g}}$. Second, a linear transformation $\mathcal{F}_L : \mathbb{R}^{\frac{N}{g}} \to \mathbb{R}^{\frac{M}{g}}$ is applied to map $\mathbf{h}^{i}$ linearly to $\mathbf{z}^{i} \in \mathbb{R}^{\frac{M}{g}}$, for each $i = \{1, \cdots, g\}$. 
The $g$ resultant output vectors $\mathbf{z}^{i}$ are concatenated to produce the final output vector $\mathbf{\bar z}  \in \mathbb{R}^M$.
\begin{equation}
\mathbf{\bar z} = \mathcal{F}_G(\mathbf{h}) = \left[\mathbf{W}^1 \cdot \mathbf{h}^1, \cdots, \mathbf{W}^g \cdot \mathbf{h}^g \right]
\label{eq:glin}
\end{equation}
GLTs learn representations at low dimensionality. Therefore, a GLT requires $g$ fewer parameters than the linear transformation. We note that GLTs are subset of linear transformations. In a linear transformation, each neuron receives an input from each element in the input vector while in a GLT, each neuron receives an input from a subset of the input vector. Therefore, GLT is the same as a linear transformation when $g=1$.

\subsection{Pyramidal Recurrent Unit}
We extend the basic gating architecture of LSTM with the pyramidal and grouped linear transformations outlined above to produce the Pyramidal Recurrent Unit (PRU), whose improved sequence modeling capacity is evidenced in Section~\ref{sec:experiments}. 

At time $t$, the PRU combines the input vector $\mathbf{x}_t$ and the previous context vector (or previous hidden state vector) $\mathbf{h}_{t-1}$ using the following transformation function as:
\begin{equation}
\mathcal{\hat G}_v(\mathbf{x}_t, \mathbf{h}_{t-1}) = \mathcal{\hat F}_P(\mathbf{x}_t) + \mathcal{F}_G(\mathbf{h}_{t-1}),
\end{equation}
where $v \in \{ f, i, c, o\}$ indexes the various gates in the LSTM model, and $\mathcal{\hat F}_P(\cdot)$ and $\mathcal{F}_G(\cdot)$ represents the pyramidal and grouped linear transformations defined in Eqns.~\ref{eq:pyr} and \ref{eq:glin}, respectively. 

We will now incorporate $\mathcal{\hat G}_v(\cdot, \cdot)$ into LSTM gating architecture to produce PRU. At time $t$, a PRU cell takes $\mathbf{x}_t \in \mathbb{R}^N$, $\mathbf{h}_{t-1} \in \mathbb{R}^M$, and $\mathbf{c}_{t-1} \in \mathbb{R}^M$ as inputs to produce forget $\mathbf{f}_t$, input $\mathbf{i}_t$, output $\mathbf{o}_t$, and content $\mathbf{\hat{c}}_t$ gate signals. The inputs are combined with these gate signals to produce context vector $\mathbf{h}_t \in \mathbb{R}^M$ and cell state $\mathbf{c}_t \in \mathbb{R}^M$. Mathematically, the PRU with the LSTM gating architecture can be defined as:
\begin{align}
\begin{split}
\mathbf{f}_t =\ & \sigma\left(\mathcal{\hat G}_f(\mathbf{x}_t, \mathbf{h}_{t-1}) \right)\\
\mathbf{i}_t =\ & \sigma\left(\mathcal{\hat G}_i(\mathbf{x}_t, \mathbf{h}_{t-1})\right)\\
\mathbf{\hat{c}}_t =\ & tanh\left(\mathcal{\hat G}_c(\mathbf{x}_t , \mathbf{h}_{t-1})\right) \\
\mathbf{o}_t =\ & \sigma\left(\mathcal{\hat G}_o(\mathbf{x}_t , \mathbf{h}_{t-1})\right)\\
\mathbf{c}_t =\ & \mathbf{f}_t \otimes \mathbf{c}_{t-1} + \mathbf{i}_t \otimes  \mathbf{\hat{c}}_t \\  \mathbf{h}_t =\ & \mathbf{o}_t \otimes tanh(\mathbf{c}_t)
\end{split}
\end{align}
\noindent where $\otimes$ represents the element-wise multiplication operation, and $\sigma$ and $tanh$ are the sigmoid and hyperbolic tangent activation functions. We note that LSTM is a special case of PRU when $g$=$K$=$1$.

\section{Experiments}
\label{sec:experiments}
To showcase the effectiveness of the PRU, we evaluate the performance on two standard datasets for word-level language modeling and compare with state-of-the-art methods. Additionally, we provide a detailed examination of the PRU and its behavior on the language modeling tasks.

\subsection{Set-up}
\label{ssec:lm}
\paragraph{Dataset:} Following recent works, we compare on two widely used datasets, the Penn Treebank (PTB) \cite{marcus1993building} as prepared by \citet{mikolov2010recurrent} and WikiText2 (WT-2) \cite{merity2016pointer}. For both datasets, we follow the same training, validation, and test splits as in \citet{merity2017regularizing}.

\paragraph{Language Model:} We extend the language model, AWD-LSTM \cite{merity2017regularizing}, by replacing LSTM layers with PRU. Our model uses  3-layers of PRU with an embedding size of 400. The number of parameters learned by state-of-the-art methods vary from 18M to 66M with majority of the methods learning about 22M to 24M parameters on the PTB dataset. For a fair comparison with state-of-the-art methods, we fix the model size to 19M and vary the value of $g$ and hidden layer sizes so that total number of learned parameters is similar across different configurations. We use 1000, 1200, and 1400 as hidden layer sizes for values of $g$=1,2, and 4, respectively. We use the same settings for the WT-2 dataset. We set the number of pyramidal levels $K$ to two in our experiments and use average pooling for sub-sampling. These values are selected based on our ablation experiments on the validation set (Section \ref{sec:abl}). We measure the performance of our models in terms of word-level perplexity. We follow the same training strategy as in \citet{merity2017regularizing}. 

To understand the effect of regularization methods on the performance of PRUs, we perform experiments under two different settings: (1) \textit{Standard dropout}: We use a standard dropout \cite{srivastava2014dropout} with probability of 0.5 after embedding layer, the output between LSTM layers, and the output of final LSTM layer. (2) \textit{Advanced dropout:} We use the same dropout techniques with the same dropout values as in \citet{merity2017regularizing}. We call this model as AWD-PRU.

\subsection{Results}
Table \ref{tab:sota} compares the performance of the PRU with state-of-the-art methods. We can see that the PRU achieves the best performance with fewer parameters. 

\begin{table*}[t!]
\centering
\resizebox{2\columnwidth}{!}{
\begin{tabular}{l|ccc|ccc}
\toprule
\multicolumn{1}{l}{} & \multicolumn{3}{|c|}{\textbf{WT-2}} & \multicolumn{3}{c}{\textbf{PTB}} \\ 
\midrule
 \textbf{Model} & \textbf{Params} & \textbf{Val} & \textbf{Test} & \textbf{Params} & \textbf{Val} & \textbf{Test} \\ 
 \midrule
Variational LSTM \cite{gal2016theoretically} & -- & --  & -- & 20 M &  -- & 78.6 \\  
CharCNN \cite{kim2016character}  &--  & --  & -- & 19 M &  -- & 78.9 \\ 
Pointer Sentinel-LSTM \cite{merity2016pointer} &--  & --  & -- & 19 M &  72.4 & 70.9 \\ 
RHN \cite{zilly2016recurrent} &--  & --  & -- & 23 M &  67.9 & 65.4 \\ 
NAS Cell \cite{zoph2016neural} &--  & --  & -- & 25 M &  -- & 64.0 \\
Variational LSTM  - \cite{inan2016tying} &  28 M & 91.5 & 87 & 24 M & 75.7 & 73.2 \\
SRU - 6 layers \cite{lei2017sru} & -- & -- & -- & 24 M & 63.4 & 60.3 \\ 
QRNN \cite{bradbury2016quasi} & -- & -- & -- & \textbf{18 M} & 82.1 & 78.3 \\ 
RAN \cite{lee2017recurrent} & -- &-- & -- & 22 M & -- & 78.5  \\
4-layer skip-connection LSTM \cite{melis2017state} &--  & --  & -- & 24 M &  60.9 & 58.3 \\
AWD-LSTM - \cite{merity2017regularizing} & 33 M & 69.1 & 66 & 24 M & 60.7 & 58.8 \\ 
AWD-LSTM - \cite{merity2017regularizing}-finetuned  & 33 M & 68.6 & 65.8 & 24 M & 60 & 57.3 \\
\midrule
Variational LSTM \cite{gal2016theoretically}  &--  & --  & -- & 66 M &  -- & 73.4 \\ 
NAS Cell \cite{zoph2016neural}  &--  & --  & -- & 54 M &  -- & 62.4 \\
Quantized LSTM - Full precision \cite{Chen2018alter} & --& --& 100.1 & --& --& 89.8 \\
Quantized LSTM - 2 bit \cite{Chen2018alter} & --&-- & 106.1 & -- & -- & 95.8 \\  
\midrule
\multicolumn{7}{c}{With standard dropout}\\
 \midrule
LSTM ($M=1000$) & 29 M & 78.93 & 75.08 & 20 M & 68.57 & 66.29  \\
LSTM ($M=1200$) & 35 M & 77.93 & 74.48 & 26 M  & 69.17 & 67.16 \\ 
LSTM ($M=1400$) & 42 M & 77.55 & 74.44 & 33 M & 70.88 &  68.55 \\ 
Ours -PRU  ($g=1$, $K=2$, $M=1000$) & 28 M & 79.15 & 76.59 & 19 M & 69.8 & 67.78 \\ 
Ours -PRU  ($g=2$, $K=2$, $M=1200$) & 28 M & 76.62 & 73.79 & 19 M & 67.17 & 64.92 \\ 
Ours -PRU ($g=4$, $K=2$, $M=1400$) & 28 M & 75.46 & 72.77 & 19 M & 64.76 & 62.42 \\
\midrule
\multicolumn{7}{c}{With advanced dropouts}\\
\midrule
Ours - AWD-PRU ($g=1$, $K=2$, $M=1000$) & 28 M &  71.84 & 68.6 & 19 M & 61.72 &  59.54\\
Ours - AWD-PRU ($g=2$, $K=2$, $M=1200$) &  28 M & 68.57 & 65.7 & 19 M & 60.81 & 58.65 \\
Ours - AWD-PRU ($g=4$, $K=2$, $M=1400$) & 28 M & 68.17 & 65.3 & 19 M & 60.62 & 58.33 \\
Ours - AWD-PRU ($g=4$, $K=2$, $M=1400$)-finetuned & \textbf{28 M} & \textbf{67.19} & \textbf{64.53} & 19 M & \textbf{58.46} & \textbf{56.56} \\ 
\bottomrule
\end{tabular}
}
\caption{Comparison of single model word-level perplexity of our model with state-of-the-art on validation and test sets of Penn Treebank and Wikitext-2 dataset. For evaluation, we select the model with minimum validation loss. Lower perplexity value represents better performance.}
\label{tab:sota} 
\end{table*}

\paragraph{Standard dropout:} PRUs achieve either the same or better performance than LSTMs. In particular, the performance of PRUs improves with the increasing value of $g$. At $g=4$, PRUs outperform LSTMs by about 4 points on the PTB dataset and by about 3 points on the WT-2 dataset. This is explained in part by the regularization effect of the grouped linear transformation (Figure \ref{fig:lstmOverHD}). With grouped linear and pyramidal transformations, PRUs learn rich representations at very high dimensional space while learning fewer parameters. On the other hand, LSTMs overfit to the training data at such high dimensions and learn $1.4\times$ to $1.8\times$ more parameters than PRUs.  

\paragraph{Advanced dropouts:} With the advanced dropouts, the performance of PRUs improves by about 4 points on the PTB dataset and 7 points on the WT-2 dataset. This further improves with finetuning on the PTB (about 2 points) and WT-2 (about 1 point) datasets. 

\paragraph{Comparison with state-of-the-art:} For similar number of parameters, the PRU with standard dropout outperforms most of the state-of-the-art methods by large margin on the PTB dataset (e.g. RAN \cite{lee2017recurrent} by 16 points with 4M less parameters, QRNN \cite{bradbury2016quasi} by 16 points with 1M more parameters, and NAS \cite{zoph2016neural} by 1.58 points with 6M less parameters). With advanced dropouts, the PRU delivers the best performance. On both datasets, the PRU improves the perplexity by about 1 point while learning 15-20\% fewer parameters. 

\paragraph{Inference:} PRU is a drop-in replacement for LSTM, therefore, it can improve language models with modern inference techniques such as dynamic evaluation \cite{krause2017dynamic}. When we evaluate PRU-based language models (only with standard dropout) with dynamic evaluation on the PTB test set, the perplexity of PRU ($g=4,k=2,M=1400$) improves from 62.42 to 55.23 while the perplexity of an LSTM ($M=1000$) with similar settings improves from 66.29 to 58.79; suggesting that modern inference techniques are equally applicable to PRU-based language models.

\subsection{Analysis} 
\label{ssec:analysis}
It is shown above that the PRU can learn representations at higher dimensionality with more generalization power, resulting in performance gains for language modeling. A closer analysis of the impact of the PRU in a language modeling system reveals several factors that help explain how the PRU achieves these gains.

\begin{figure}[t!]
\centering
\includegraphics[width=\columnwidth]{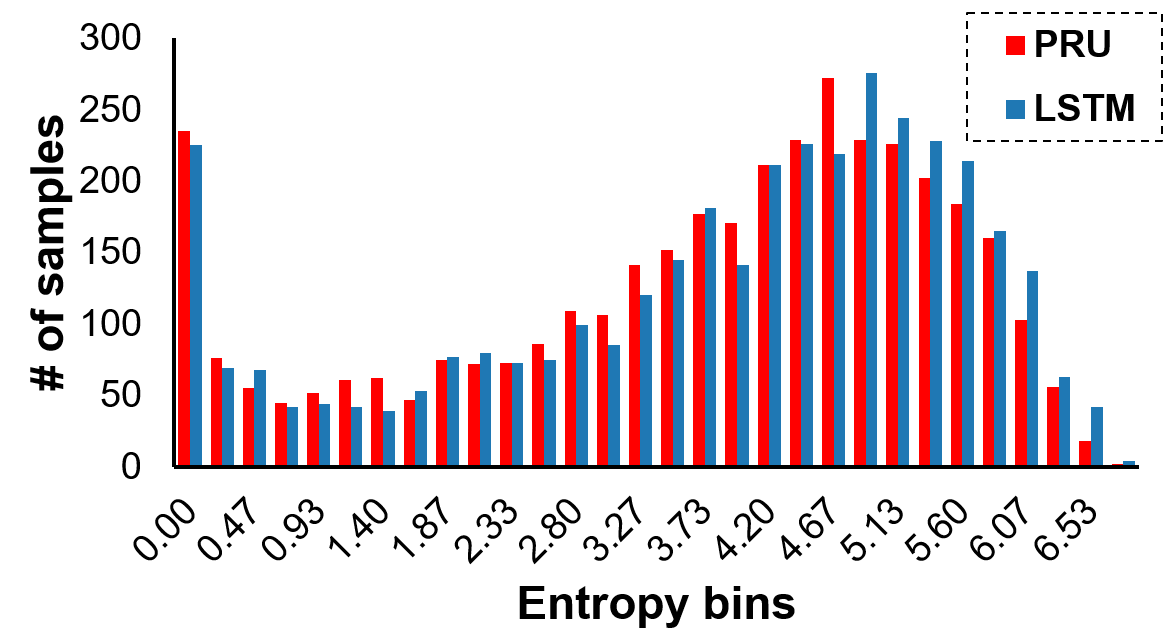}
\caption{Histogram of the entropies of next-token distributions predicted by the PRU (mean 3.80) and the LSTM (mean 3.93) on the PTB validation set. Lower entropy values indicate higher confidence decisions, which is desirable if decisions are often correct.}
\label{fig:ent}
\end{figure}

\paragraph{Confidence:} As exemplified in Table~\ref{tab:heatmaps}, the PRU tends toward more confident decisions, placing more of the probability mass on the top next-word prediction than the LSTM. To quantify this effect, we calculate the entropy of the next-token distribution for both the PRU and the LSTM using 3687 contexts from the PTB validation set. Figure~\ref{fig:ent} shows a histogram of the entropies of the distribution, where bins of size 0.23 are used to effect categories. We see that the PRU more often produces lower entropy distributions corresponding to higher confidences for next-token choices. This is evidenced by the mass of the red PRU curve lying in the lower entropy ranges compared to the blue LSTM's curve. The PRU can produce confident decisions in part because more information is encoded in the higher dimensional context vectors. 

\begin{figure}[t!]
\centering
\includegraphics[width=\columnwidth]{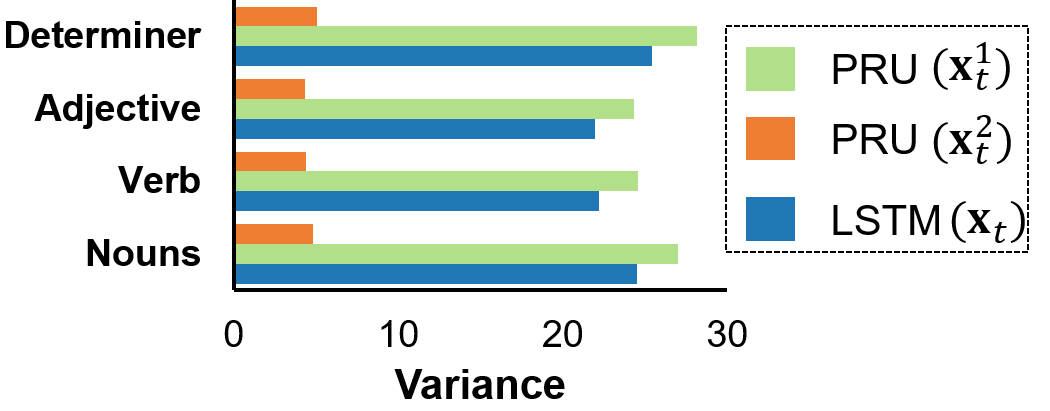}
\caption{Variance of learned word embeddings for different categories of words on the PTB validation set. We compute the variance of a group of embeddings as the average squared euclidean distance to their mean. Higher variance may allow for better intra-category distinctions. The PRU with pyramid levels 1 and 2 is shown.}
\label{fig:var}
\end{figure}

\paragraph{Variance in word embeddings:} 
The PRU has the ability to model individual words at different resolutions through the pyramidal transform; which provides multiple paths for the gradient to the embedding layer (similar to multi-task learning) and improves the flow of information. When considering the embeddings by part of speech, we find that the pyramid level 1 embeddings exhibit higher variance than the LSTM across all POS categories (Figure~\ref{fig:var}), and that pyramid level 2 embeddings show extremely low variance\footnote{POS categories are computed using NLTK toolkit.}. We hypothesize that the LSTM must encode both coarse group similarities and individual word differences into the same vector space, reducing the space between individual words of the same category. The PRU can rely on the subsampled embeddings to account for coarse-grained group similarities, allowing for finer individual word distinctions in the embedding layer. This hypothesis is strengthened by the entropy results described above: a model which can make finer distinctions between individual words can more confidently assign probability mass. A model that cannot make these distinctions, such as the LSTM, must spread its probability mass across a larger class of similar words. 
%
\begin{table*}[t!]
\centering
\begin{subtable}{1.9\columnwidth}
\resizebox{\columnwidth}{!}{
\begin{tabular}{l|ll}
\toprule
\multicolumn{1}{c|}{\bf Gradient-based sensitivity analysis heatmaps} & \multicolumn{1}{c}{\bf LSTM top-5} & \multicolumn{1}{c}{\bf PRU top-5}\\
\midrule
\includegraphics[height=40px]{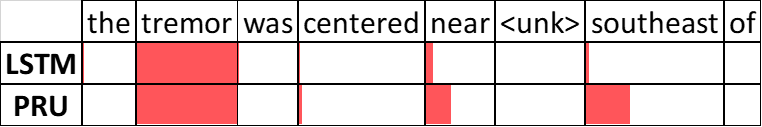} & \includegraphics[height=50px]{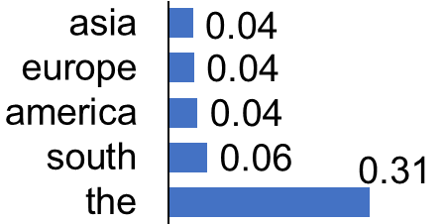} & \includegraphics[height=50px]{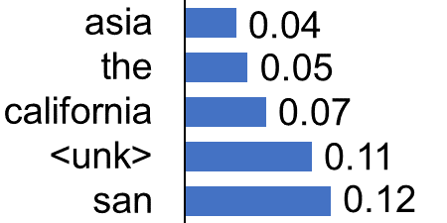}\\
\multicolumn{1}{l}{\textbf{Reference:} the tremor was centered near $<$unk$>$ southeast of \textbf{san} francisco} & & \\
\midrule
\includegraphics[height=40px]{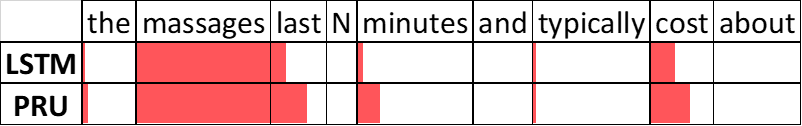} & \includegraphics[height=50px]{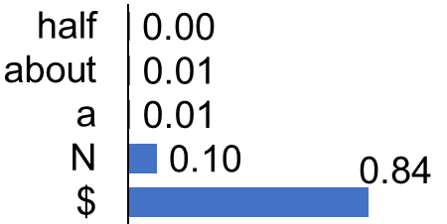} & \includegraphics[height=50px]{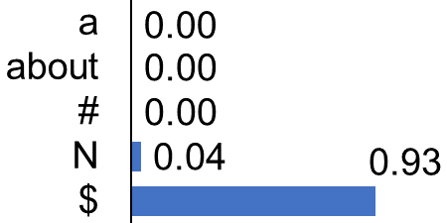}\\
\multicolumn{3}{l}{\textbf{Reference:} the massages last N minutes and typically cost about \textbf{\$} N.}\\
\midrule
\includegraphics[height=40px]{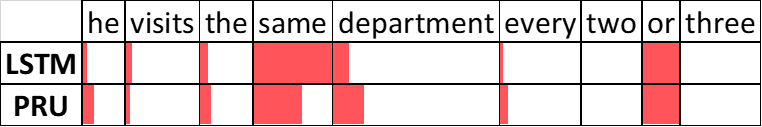} & \includegraphics[height=50px]{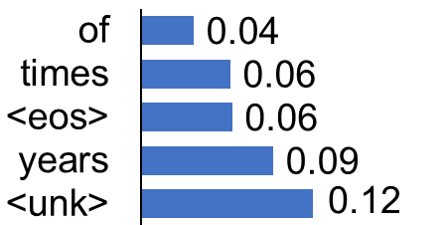} & \includegraphics[height=50px]{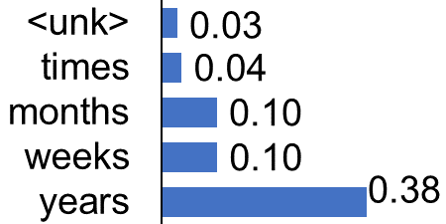}\\
\multicolumn{3}{l}{\textbf{Reference:} he visits the same department every two or three \textbf{weeks}.}\\
\midrule
    	\includegraphics[height=40px]{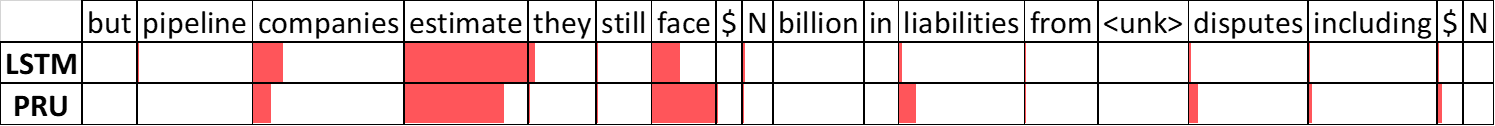}&
    	\includegraphics[height=50px]{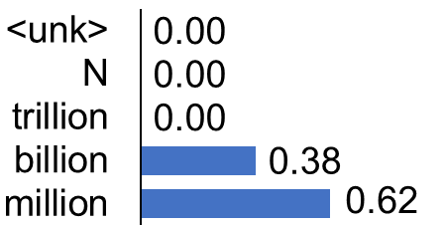}&
    	\includegraphics[height=50px]{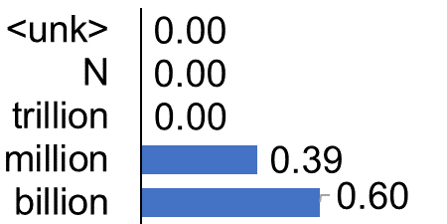}\\
\multicolumn{3}{l}{\textbf{Reference:}   but pipeline companies estimate they still face \$ N billion in liabilities from $<$unk$>$ disputes including \$ N \textbf{billion}.}\\
\midrule
    	\includegraphics[height=40px]{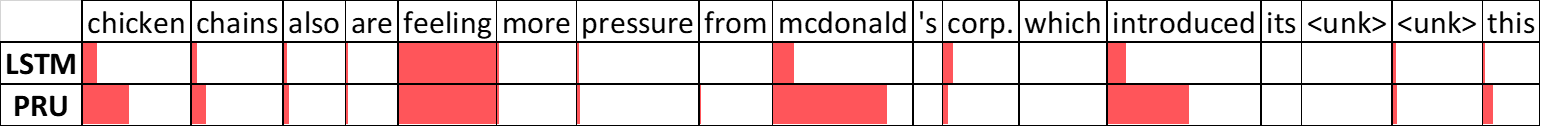}&
    	\includegraphics[height=50px]{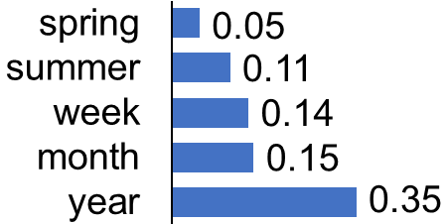}&
    	\includegraphics[height=50px]{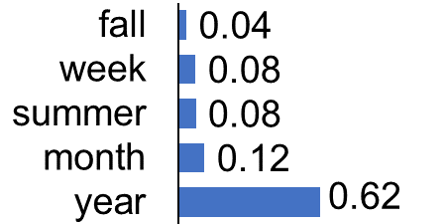}\\
\multicolumn{3}{l}{\textbf{Reference:}  chicken chains also are feeling more pressure from mcdonald's corp. which introduced its $<$unk$>$ $<$unk$>$ this \textbf{year}.} \\
\bottomrule
\end{tabular}
}
\caption{Gradient-based saliency analysis. Salience score is proportional to cell coverage in \textcolor{red}{\bf red}.}
\label{tab:heatmaps}
\end{subtable}
\begin{subtable}{2\columnwidth}
\resizebox{\columnwidth}{!}{
\begin{tabular}{lcccc}
\rotatebox{90}{\LARGE LSTM} & \raisebox{-.5\height}{\includegraphics[width=0.65\columnwidth]{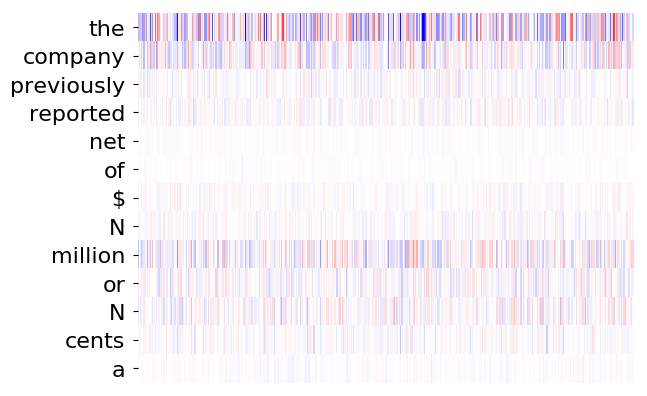}} & \raisebox{-.5\height}{\includegraphics[width=0.65\columnwidth]{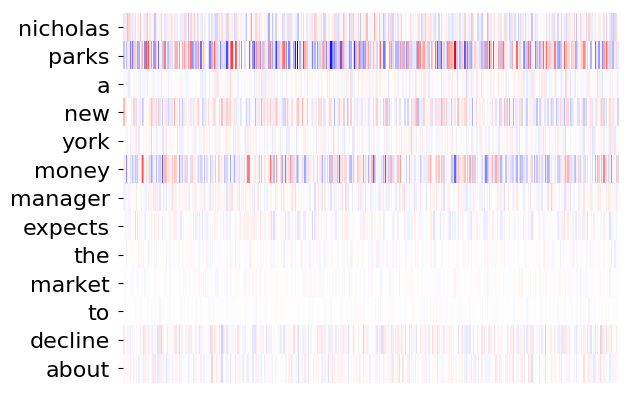}} & \raisebox{-.5\height}{\includegraphics[width=0.65\columnwidth]{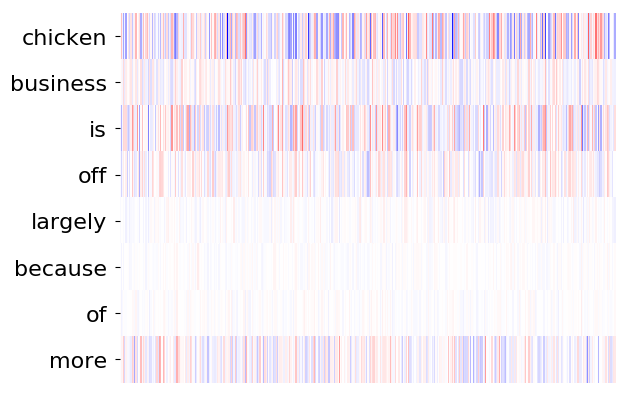}} & \raisebox{-.5\height}{\includegraphics[width=0.1\columnwidth]{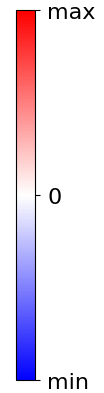}}  \\ 
\rotatebox{90}{\LARGE PRU} & \raisebox{-.5\height}{\includegraphics[width=0.65\columnwidth]{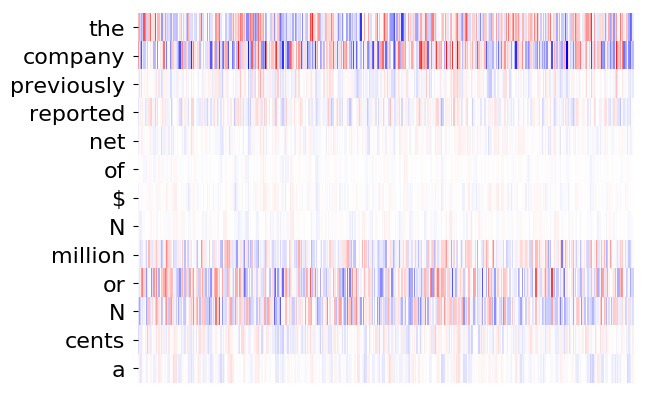}} & \raisebox{-.5\height}{\includegraphics[width=0.65\columnwidth]{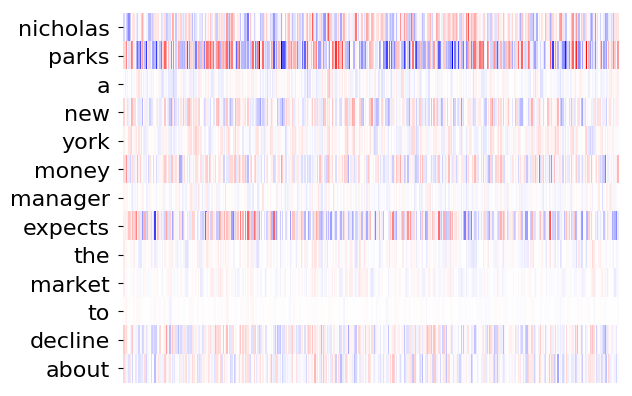}} & \raisebox{-.5\height}{\includegraphics[width=0.65\columnwidth]{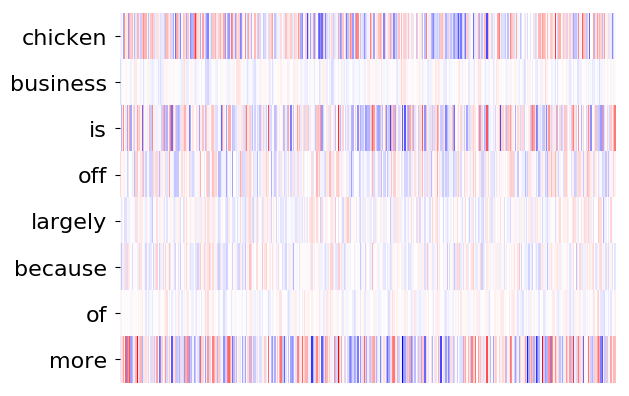}} &  \\ 
\end{tabular}
l}
\caption{Gradients during back-propagation for a test sequence (x-axis: dimensions of word vector, y-axis: test sequence)}
\label{tab:gradientAna}
\end{subtable}
\caption{Qualitative comparison between the LSTM and the PRU: (a) Gradient-based saliency analysis along with top-5 predicted words. (b) Gradients during back-propagation. For computing the gradients for a given test sequence, the top-1 predicted word was used as the \textit{true} predicted word. Best viewed in color.} 
\end{table*}
%
\begin{figure*}[t!]
\centering
\begin{subfigure}[b]{0.95\columnwidth}
\centering
\includegraphics[width=\columnwidth]{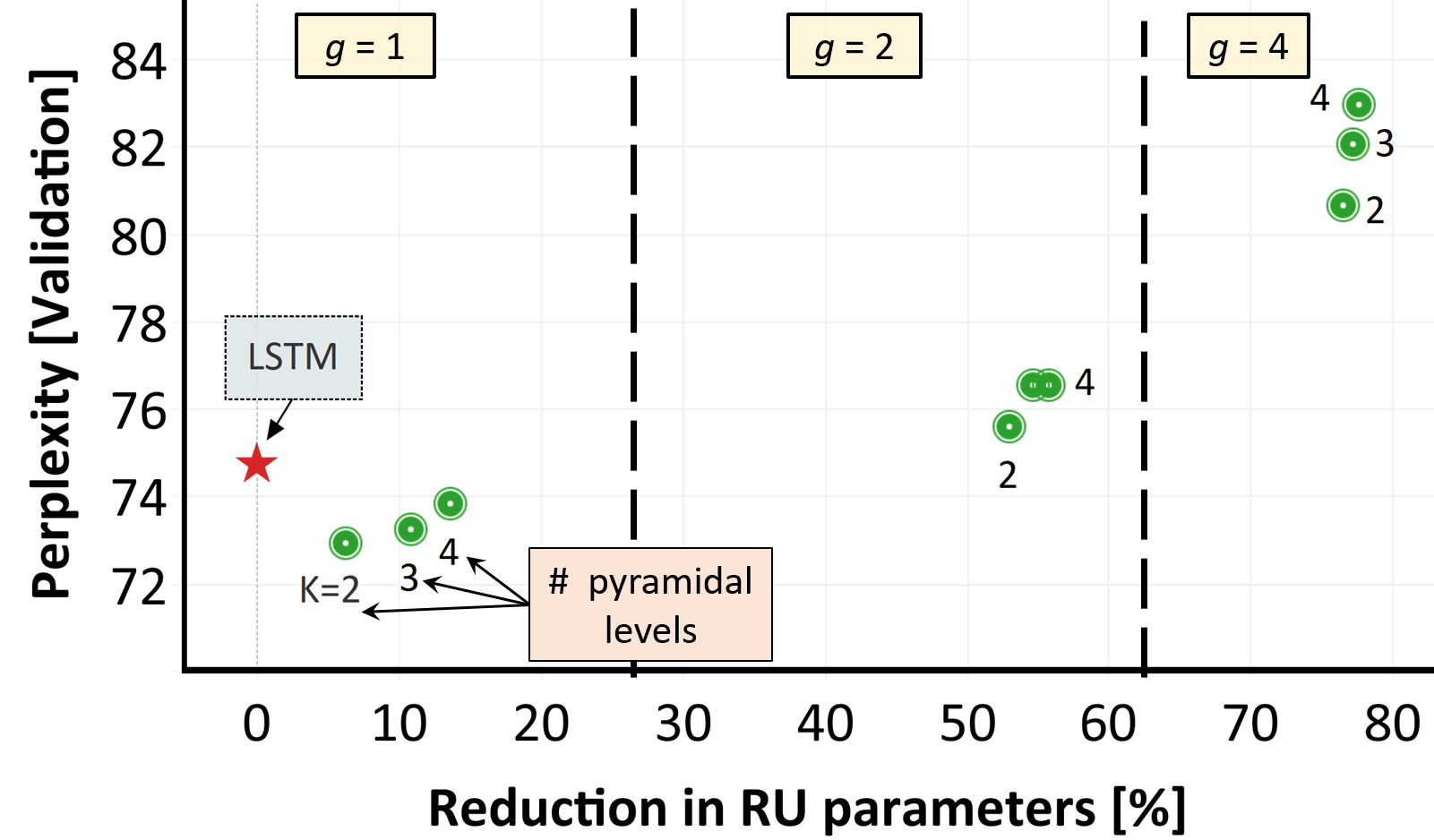}
\caption{PTB}
\end{subfigure}
\hfill
\begin{subfigure}[b]{0.95\columnwidth}
\centering
\includegraphics[width=\columnwidth]{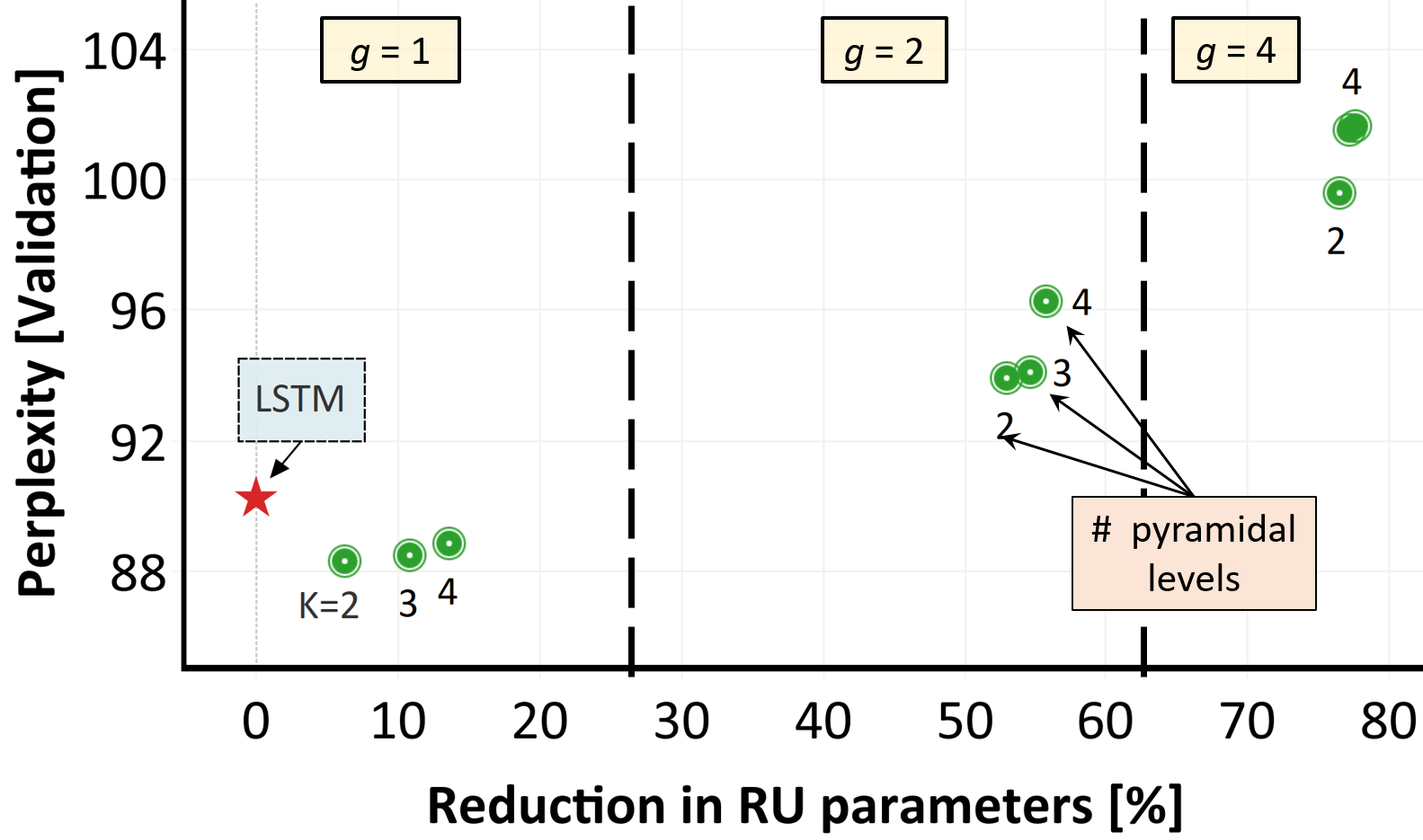}
\caption{WT-2}
\end{subfigure}
\caption{Impact of number of groups $g$ and pyramidal levels $K$ on the perplexity. Reduction in recurrent unit (RU) parameters is computed with respect to LSTM. Lower perplexity value represents better performance.}
\label{fig:impactK}
\end{figure*}
\paragraph{Gradient-based analysis:} Saliency analysis using gradients help identify relevant words in a test sequence that contribute to the prediction \cite{gevrey2003review, li2015visualizing, arras2017explaining}. These approaches compute the relevance as the squared norm of the gradients obtained through back-propagation. Table~\ref{tab:heatmaps} visualizes the heatmaps for different sequences. PRUs, in general, give more relevance to contextual words than LSTMs, such as southeast (sample 1), cost (sample 2), face (sample 4), and introduced (sample 5), which help in making more confident decisions. Furthermore, when gradients during back-propagation are visualized \cite{selvaraju2016grad} (Table \ref{tab:gradientAna}), we find that PRUs have better gradient coverage than LSTMs, suggesting PRUs use more features than LSTMs that contributes to the decision. This also suggests that PRUs update more parameters at each iteration which results in faster training. Language model in \cite{merity2017regularizing} takes 500 and 750 epochs to converge with PRU and LSTM as a recurrent unit, respectively.

\subsection{Ablation studies}
\label{sec:abl}
In this section, we provide a systematic analysis of our design choices. 
Our training methodology is the same as described in Section \ref{ssec:lm} with the standard dropouts. For a thorough understanding of our design choices, we use a language model with a single layer of PRU and fix the size of embedding and hidden layers to 600. The word-level perplexities are reported on the validation sets of the PTB and the WT-2 datasets.

\paragraph{Pyramidal levels $K$ and groups $g$:} The two hyper-parameters  that control the trade-off between performance and number of parameters in PRUs are the number of pyramidal levels $K$ and groups $g$. Figure \ref{fig:impactK} provides a trade-off between perplexity and recurrent unit (RU) parameters\footnote{\# total params = \# embedding params + \# RU params}. 

\textit{Variable $K$ and fixed $g$}: When we increase the number of pyramidal levels $K$ at a fixed value of $g$, the performance of the PRU drops by about 1 to 4 points while reducing the total number of recurrent unit parameters by up to 15\%. We note that the PRU with $K=4$ at $g=1$ delivers similar performance as the LSTM while learning about 15\% fewer recurrent unit parameters.   

\textit{Fixed $K$ and variable $g$}: When we vary the value of $g$ at fixed number of pyramidal levels $K$, the total number of recurrent unit parameters decreases significantly with a minimal impact on the perplexity. For example, PRUs with $K=2$ and $g=4$ learns 77\% fewer recurrent unit parameters while its perplexity (lower is better) increases by about 12\% in comparison to LSTMs. Moreover, the decrease in number of parameters at higher value of $g$ enables PRUs to learn the representations in high dimensional space with better generalizability (Table \ref{tab:sota}).

\paragraph{Transformations:} Table \ref{tab:diffTransformations} shows the impact of different transformations of the input vector $\mathbf{x}_t$ and the context vector $\mathbf{h}_{t-1}$. We make following observations: (1) Using the pyramidal transformation for the input vectors improves the perplexity by about 1 point on both the PTB and WT-2 datasets while reducing the number of recurrent unit parameters by about 14\% (see R1 and R4). We note that the performance of the PRU drops by up to 1 point when residual connections are not used (R4 and R6). (2) Using the grouped linear transformation for context vectors reduces the total number of recurrent unit parameters by about 75\% while the performance drops by about 11\% (see R3 and R4). When we use the pyramidal transformation instead of the  linear transformation, the performance drops by up to 2\% while there is no significant drop in the number of parameters (R4 and R5).

\begin{table}[b!]
\resizebox{\columnwidth}{!}{
\begin{tabular}{c|cc|cc|cc}
\toprule
 & & & \multicolumn{2}{c}{\textbf{PTB}} & \multicolumn{2}{|c}{\textbf{WT-2}} \\
 & \multicolumn{2}{c|}{\textbf{Transformations}} & \textbf{PPL} & \textbf{\# Params} & \textbf{PPL} & \textbf{\# Params} \\
& Context & Input & & (total/RU) &  & (total/RU) \\
\midrule
R1 & LT & LT & 74.80 & 8.8/2.9 & 89.30 & 22.8/2.9\\
R2& GLT & GLT & 84.38 & \textbf{6.5/0.5} & 104.13 & \textbf{20.46/0.5} \\
R3 & GLT & PT & 82.67 & 6.6/0.64 & 99.57 & 20.6/0.64 \\
R4 & LT & PT & \textbf{74.18} & 8.5/2.5 & \textbf{88.31} & 22.5/2.5\\
R5 & PT & PT & 75.80 & 8.1/2.1 & 90.56 & 22.1/2.1\\
\midrule
R6 & LT & PT$^\dagger$ & 75.61 & 8.5/2.5 & 89.27 & 22.5/2.5\\
\bottomrule
\end{tabular}
}
\caption{Impact of different transformations used for processing input and context vectors (LT - linear transformation, PT - pyramidal transformation, and GLT - grouped linear transformation). Here, $^\dagger$ represents that PT was used without residual connection, PPL represents word-level perplexity (lower is better), and the number of parameters are in million. We used $K$=$g$=$4$ in our experiments.}
\label{tab:diffTransformations}
\end{table}

\paragraph{Subsampling:} 
We set sub-sampling kernel $\kappa$ (Eq. \ref{eq:kern}) with stride $s=2$ and size of 3 ($e=1$) in four different ways: (1) \textit{Skip:} We skip every other element in the input vector. (2) \textit{Convolution:} We initialize the elements of $\kappa$ randomly from normal distribution and learn them during training the model. We limit the output values between -1 and 1 using $tanh$ activation function to make training stable. (3) \textit{Avg. pool:} We initialize the elements of $\kappa$ to $\frac{1}{3}$. (4) \textit{Max pool:} We select the maximum value in the kernel window $\kappa$.

\begin{table}[t!]
\centering
\begin{small}
\resizebox{\columnwidth}{!}{
\begin{tabular}{l|cccc}
\toprule
\textbf{Dataset} & \textbf{Skip} & \textbf{Max pool} & \textbf{Avg. Pool} & \textbf{Convolution} \\
\midrule
\textbf{PTB} & 75.12 & 87.6 & \textbf{73.86} & 81.56 \\
\textbf{WT-2} & 89.24 & 107.63 & \textbf{88.88} & 93.16 \\
\hline
\end{tabular}
}
\end{small}
\caption{Impact of different sub-sampling methods on the word-level perplexity (lower is better). We used $g$=$1$ and $K$=$4$ in our experiments.}
\label{tab:sampleMeth}
\end{table}

Table \ref{tab:sampleMeth} compares the performance of the PRU with different sampling methods. Average pooling performs the best while skipping give comparable performance. Both of these methods enable the network to learn richer word representations while representing the input vector in different forms, thus delivering higher performance. Surprisingly, a convolution-based sub-sampling method does not perform as well as the averaging method. The $tanh$ function used after convolution limits the range of output values which are further limited by the LSTM gating structure, thereby impeding in the flow of information inside the cell. Max pooling forces the network to learn representations from high magnitude elements, thus distinguishing features between elements vanishes, resulting in poor performance.

\section{Conclusion}
We introduce the Pyramidal Recurrent Unit, which better model contextual information by admitting higher dimensional representations with good generalizability. When applied to the task of language modeling, PRUs improve perplexity across several settings, including recent state-of-the-art systems. Our analysis shows that the PRU improves the flow of gradient and expand the word embedding subspace, resulting in more confident decisions. Here we have shown improvements for language modeling. In future, we plan to study the performance of PRUs on different tasks, including machine translation and question answering. In addition, we will study the performance of the PRU on language modeling with more recent inference techniques, such as dynamic evaluation and mixture of softmax. 

\subsection*{Acknowledgments}
This research was supported by  NSF (IIS 1616112, III 1703166), Allen Distinguished Investigator Award, and gifts from Allen Institute for AI, Google, Amazon, and Bloomberg.  We are grateful to Aaron Jaech, Hannah Rashkin, Mandar Joshi, Aniruddha Kembhavi, and anonymous reviewers for their helpful comments.

\bibliography{emnlp2018}
\bibliographystyle{acl_natbib_nourl}

\end{document}